\documentclass[runningheads]{llncs}
\usepackage[T1]{fontenc}
\usepackage{graphicx}
\usepackage{booktabs}
\usepackage[misc]{ifsym}
\usepackage{algorithm}
\usepackage{float}
\usepackage{algorithmic}
\usepackage{amsmath} % 用于数学符号
\usepackage{amssymb} % 用于数学字体
\usepackage{booktabs}  % 必须：用于绘制专业的三线表
\usepackage{multirow}  % 用于处理复杂的表格布局
\usepackage{graphicx}  % 用于 resizebox 缩放表格
\usepackage{subcaption}
\usepackage{enumitem}

\usepackage{mwe}
\begin{document}

\title{GLIDE: Graph-guided Leap Inference for Diffusion Estimation of Spatio-Temporal Point Processes}

\titlerunning{GLIDE: Graph-guided Leap Inference for STPPs}

\author{
Guanyu Zhou \and
Yao Liu \and
Yanglei Gan \and
Yuxiang Cai \and
Peng He \and
Run Lin \and
Yuxiang Liu \and
Qiao Liu
}

\institute{
University of Electronic Science and Technology of China, Chengdu, China \\
\email{
guanyuzhou@std.uestc.edu.cn,
liuyao@uestc.edu.cn,
yangleigan@std.uestc.edu.cn,
yuxiangcai@std.uestc.edu.cn,
hepenglk@std.uestc.edu.cn,
runlin@std.uestc.edu.cn,
2023210812@std.uestc.edu.cn,
qliu@uestc.edu.cn
}
}

\maketitle
\begin{abstract}
Spatio-temporal point processes (STPPs) provide a principled framework for modeling asynchronous events in continuous time and space. Recent diffusion-based approaches offer a flexible alternative to deterministic prediction by modeling complex conditional distributions, but their application to STPPs remains challenging: reverse sampling from pure noise is costly, and weak structural constraints in sparse spatial domains can lead to poorly localized probability mass. We propose \textbf{GLIDE} (Graph-guided Leap Inference for Diffusion Estimation), a conditional diffusion framework for next-event modeling in STPPs. GLIDE organizes historical events into a multi-scale historical graph and encodes temporal evolution and spatial topology through a dual-stream architecture, yielding a structured conditioning context for a dual-branch diffusion denoiser. It further introduces a prior-guided leap inference mechanism, in which a lightweight mean predictor provides a deterministic anchor and the reverse process starts from an intermediate diffusion step instead of from pure Gaussian noise. Experiments on multiple real-world datasets show that GLIDE improves both distribution fitting and next-event prediction, with the largest gains appearing on the spatial side. The results also indicate that prior-guided leap inference substantially reduces reverse-sampling cost while preserving the stochastic generation capability of diffusion models. The code is available at \url{https://github.com/AONE-NLP/GLIDE}.

\end{abstract}

\section{Introduction}

Spatio-Temporal Point Processes (STPPs) provide a probabilistic framework for modeling asynchronous events in continuous time and space, and have been widely used in applications such as earthquake forecasting, epidemic modeling, urban mobility analysis, and crime prediction \cite{hawkes1971spectra,ogata1988statistical}. The core goal is to estimate the conditional distribution of the next event given the observed history, i.e., when and where the next event will occur.

\begin{figure}[t]
    \centering
    \includegraphics[width=\linewidth]{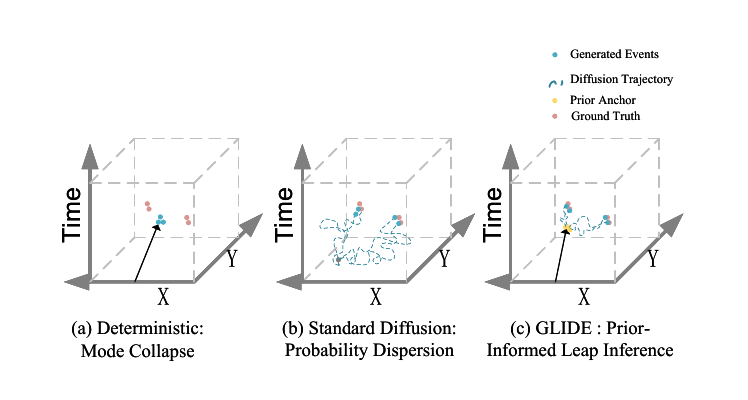}
    \caption{Comparison of STPP generative paradigms. Compared to (a) deterministic mode collapse and (b) unconstrained probability dispersion in standard diffusion, (c) GLIDE leverages a prior-informed anchor for rapid and precise leap inference.}
    \label{fig:teaser}
\end{figure}
Despite recent progress, existing deep STPP models still face a fundamental trade-off between accurate point prediction and expressive distribution modeling. Deterministic neural point process models, including recurrent and attention-based variants, can capture sequential dependence effectively, but often exhibit mean-seeking behavior when the future spatial distribution is highly multi-modal \cite{du2016recurrent,mei2017neural,zuo2020transformer,zhang2020self}. As a result, predictions may collapse toward spatial centroids instead of reflecting true event hotspots.

Generative modeling offers a more flexible alternative. Diffusion and score-based models are powerful tools for representing complex probability densities \cite{ho2020denoising,song2021score}, and have recently been introduced to STPP modeling \cite{yuan2023dstpp,li2024beyond}. However, directly applying standard diffusion to sparse continuous spatio-temporal domains remains inefficient and weakly constrained: reverse generation from pure Gaussian noise requires many denoising steps, and the lack of explicit structural guidance may spread probability mass into invalid spatial regions.

To address these issues, we propose \textbf{GLIDE} (Graph-guided Leap Inference for Diffusion Estimation), a unified framework that combines graph-guided conditioning with efficient diffusion sampling, as illustrated in Fig.~\ref{fig:teaser}. GLIDE constructs a multi-scale dynamic correlation graph from historical events and encodes temporal evolution and spatial structure using a dual-stream architecture. We further introduce a \textbf{Prior-Informed Leap Inference} strategy, which first predicts a deterministic anchor for the target event and then starts reverse diffusion from an intermediate step around this anchor. This turns generation from a costly global search into a local refinement process, improving both efficiency and spatial fidelity.

Our contributions are summarized as follows:
\begin{itemize}[leftmargin=*]
\item \textbf{A graph-guided diffusion framework for STPP modeling.} We propose GLIDE, which integrates multi-scale dynamic correlation graphs with a dual-stream diffusion architecture to model complex spatio-temporal dependencies.
\item \textbf{Prior-informed leap inference for efficient sampling.} We introduce a leap sampling strategy that anchors the diffusion trajectory with a deterministic prior, substantially reducing inference cost while preserving stochastic generation.
\item \textbf{Strong empirical performance.} Experiments on multiple real-world datasets show that GLIDE achieves state-of-the-art performance in distribution fitting and event prediction, while providing up to $3.1\times$ faster inference than standard diffusion baselines.
\end{itemize}

\section{Related Work}

\paragraph{Deep Spatio-Temporal Point Processes.}
Classical STPPs rely on hand-crafted conditional intensity functions, such as Hawkes-type self-exciting processes and earthquake-oriented statistical models \cite{hawkes1971spectra,ogata1988statistical}. Notably, these explicitly parameterized point processes have also been successfully adapted to model localized social behaviors, such as the contagion of crime and epidemic outbreaks \cite{mohler2011self}.
To improve expressiveness, neural point process models parameterize event dynamics with recurrent or attention-based architectures, including RMTPP, Neural Hawkes Process, Transformer Hawkes Process, and SAHP \cite{du2016recurrent,mei2017neural,zuo2020transformer,zhang2020self}. 
Other notable intensity-free and fully neural approaches further improve flexibility without restricting the functional form \cite{shchur2020intensity,omi2019fully}.
For continuous spatio-temporal event modeling, Neural Spatio-Temporal Point Processes leverage Neural ODEs and continuous-time normalizing flows, while DeepSTPP introduces a latent spatio-temporal intensity with amortized variational inference \cite{chen2021nstpp,zhou2022deepstpp}.
Although these methods improve flexibility, they still mainly rely on intensity-based likelihood modeling and can struggle with highly multi-modal spatial distributions.

\paragraph{Spatio-Temporal Graph \& Sequence Modeling.}
Beyond asynchronous point processes, the modeling of structured time-series and macroscopic traffic often relies on sequence and graph encoders. Since the introduction of the self-attention mechanism \cite{vaswani2017attention}, Transformer-based architectures have dominated long-sequence forecasting \cite{zhou2021informer}. Concurrently, built upon foundational Graph Convolutional Networks (GCNs) and Graph Attention Networks (GATs) \cite{kipf2016semi,velickovic2018graph}, spatial-temporal graph neural networks (STGNNs) such as DCRNN \cite{li2018diffusion}, ASTGCN \cite{guo2019attention}, and Graph WaveNet \cite{wu2019graph}, have proven highly effective in modeling complex mobility. GLIDE takes inspiration from these explicit topological and sequential priors but significantly extends them from deterministic forecasting into a generative continuous-time diffusion framework.

\paragraph{Generative Modeling for STPPs.}
Generative approaches seek to model the full conditional distribution of future events rather than only point estimates.
Neural ODE/CNF-based STPP models offer expressive density modeling but require solving continuous-time dynamics during training or inference \cite{chen2021nstpp}.
Concurrently, fueled by the success of Latent Diffusion Models and robust conditioning techniques \cite{rombach2022high,ho2022classifier}, diffusion models have demonstrated remarkable success in broader spatio-temporal forecasting and time series imputation tasks \cite{rasul2021autoregressive,tashiro2021csdi,wen2023diffstg}.
Specifically for point processes, DSTPP introduces diffusion modeling for joint spatio-temporal event generation and demonstrates the benefit of diffusion for STPPs \cite{yuan2023dstpp}.
More recently, SMASH adopts a score-matching pseudolikelihood objective to avoid intractable normalization and supports uncertainty quantification \cite{li2024beyond}.

\paragraph{Accelerated Diffusion Inference.}
To overcome the high latency of generative sampling, non-Markovian sampling like DDIM \cite{song2021ddim} has become a standard practice. Moreover, truncated diffusion and early-stopping mechanisms \cite{mao2023leapfrog,lyu2022accelerating} have been actively explored. However, existing trajectory truncation methods like Leapfrog Diffusion (LED) \cite{mao2023leapfrog} often rely on continuous Gaussian Mixture Models (GMMs) for initialization, which are ill-suited for the highly sparse and discrete nature of asynchronous STPPs. In contrast, GLIDE introduces a prior-informed leap inference strategy built around a deterministic anchor derived from graph-conditioned history, improving both sampling efficiency and spatial fidelity.

\section{Preliminaries}

\subsection{Problem Formulation}

A spatio-temporal point process (STPP) is a sequence of asynchronous events
$\mathcal{S}=\{e_i\}_{i=1}^{N}$, where each event is represented as
$e_i=(t_i,\mathbf{s}_i)$ with $t_1<t_2<\cdots<t_N$ and
$\mathbf{s}_i \in \mathcal{X}\subset\mathbb{R}^{d}$.
Given the observed history
$\mathcal{H}_N=\{(t_i,\mathbf{s}_i)\}_{i=1}^{N}$,
the goal is to model the conditional distribution of the next event.

In our formulation, the temporal target is the next inter-event time
$\Delta t_{N+1}=t_{N+1}-t_N$ rather than the absolute timestamp.
Therefore, we represent the prediction target as
$\mathbf{x}_0=[\Delta t_{N+1}, \mathbf{s}_{N+1}] \in \mathbb{R}^{1+d}$,
and aim to learn the conditional density
$p(\mathbf{x}_0 \mid \mathcal{H}_N)$.

\subsection{Conditional Diffusion for STPPs}

Following denoising diffusion probabilistic models, we model
$p(\mathbf{x}_0 \mid \mathcal{H}_N)$ through a forward noising process and a
learned reverse denoising process \cite{ho2020denoising,song2021score}.
Given a variance schedule $\{\beta_k\}_{k=1}^{K}$, the forward process is
defined as:
\begin{equation}
q(\mathbf{x}_k \mid \mathbf{x}_{k-1}) =
\mathcal{N}\!\left(\sqrt{1-\beta_k}\mathbf{x}_{k-1}, \beta_k \mathbf{I}\right),
\end{equation}
with the closed-form marginal:
\begin{equation}
q(\mathbf{x}_k \mid \mathbf{x}_0) =
\mathcal{N}\!\left(\sqrt{\bar{\alpha}_k}\mathbf{x}_0,
(1-\bar{\alpha}_k)\mathbf{I}\right),
\end{equation}
where $\alpha_k=1-\beta_k$ and
$\bar{\alpha}_k=\prod_{j=1}^{k}\alpha_j$.

The reverse model is parameterized as:
\begin{equation}
p_{\theta}(\mathbf{x}_{k-1}\mid \mathbf{x}_k,\mathcal{H}_N)
=
\mathcal{N}\!\left(
\boldsymbol{\mu}_{\theta}(\mathbf{x}_k,k,\mathcal{H}_N),
\mathbf{\Sigma}_k
\right),
\end{equation}
which is trained with the standard noise-prediction objective:
\begin{equation}
\mathcal{L}_{\mathrm{diff}}
=
\mathbb{E}_{\mathbf{x}_0,\boldsymbol{\epsilon},k}
\left[
\left\|
\boldsymbol{\epsilon}
-
\boldsymbol{\epsilon}_{\theta}(\mathbf{x}_k,k,\mathcal{H}_N)
\right\|_2^2
\right].
\end{equation}

For sparse STPP domains, however, starting reverse generation from pure
Gaussian noise is often inefficient because the valid probability mass is
concentrated in a small subset of the spatio-temporal space. This observation
motivates the prior-informed leap inference strategy introduced in
Section~\ref{sec:leap_inference}.

\begin{figure}[t]
  \centering
  \includegraphics[width=\linewidth]{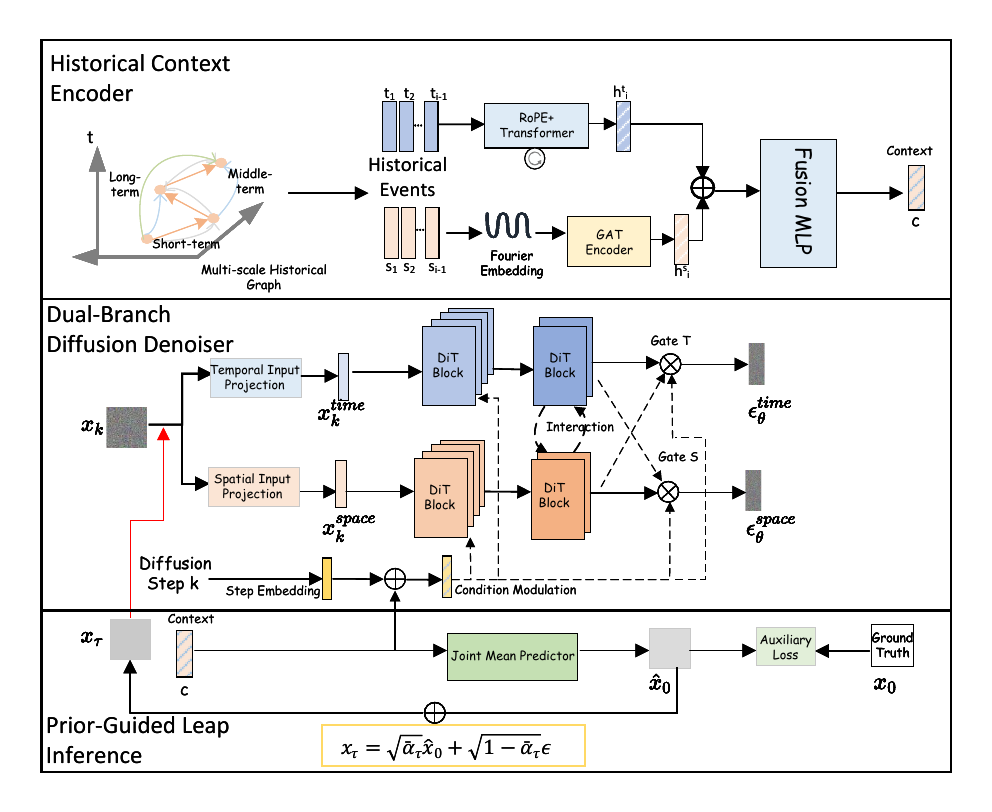} % Replace with your filename
  \caption{The overall workflow of GLIDE.}
  \label{fig:overall_workflow}
\end{figure}

\section{Methodology}
\label{sec:method}

GLIDE models the next-event distribution by coupling graph-structured historical encoding with conditional diffusion. Historical events are organized into a multi-scale historical graph. Temporal evolution and spatial topology are encoded by two separate streams, whose fused representation conditions a dual-branch Diffusion Transformer. A lightweight mean predictor provides a deterministic anchor that shortens the reverse trajectory during inference.

\subsection{Multi-Scale Dynamic Graph Construction}

Let the observed history be $\mathcal{H}_N=\{(t_i,s_i)\}_{i=1}^{N}$, where \(t_i\) is the timestamp and \(s_i\in\mathbb{R}^d\) is the spatial location. The history is represented as a directed graph $\mathcal{G}=(\mathcal{V},\mathcal{E})$ with one node per historical event. Edges are constructed under a set of historical index windows $\mathcal{K}_{\mathrm{scales}}=\{k_1,\ldots,k_m\}$.

For each target node and each scale \(k\in\mathcal{K}_{\mathrm{scales}}\), a small set of spatially close historical events is selected from the corresponding historical window and connected by directed edges. This preserves temporal order while retaining both recent local interactions and longer-range dependencies.

For an edge from event \(j\) to event \(i\), let \(\Delta t_{ji}\) denote the normalized time difference and \(\Delta s_{ji}\) denote the Euclidean spatial distance. The edge feature is defined as:
\begin{equation}
e_{ji} = [\Delta t_{ji}\cdot \delta_t,\; \Delta s_{ji}\cdot \delta_s,\; w_{\mathrm{scale}}],
\end{equation}
where \(w_{\mathrm{scale}}\) identifies the scale of the corresponding window. Following the decay intuition of self-exciting point processes \cite{hawkes1971spectra}, the temporal and spatial factors are written as:
\begin{equation}
\delta_t = \sigma\!\left(\kappa \cdot \frac{\Delta t_{ji}}{T_{\mathrm{span}}}\right), 
\qquad
\delta_s = \lambda_{\mathrm{boost}} \cdot \exp(-|\Delta s_{ji}|),
\end{equation}
where \(T_{\mathrm{span}}\) is the global time span, \(\sigma(\cdot)\) is the sigmoid function, and \(\kappa,\lambda_{\mathrm{boost}}\) are hyperparameters.

Node features combine normalized event attributes with multi-scale local summaries. For each event, temporal statistics are extracted as $f_{\mathrm{temp}}(w)=[\mu_{\Delta t},\sigma_{\Delta t},\mathrm{Trend}_{\Delta t}]$, and spatial statistics are computed analogously as $f_{\mathrm{spat}}(w)=[\mu_s,\sigma_s,v_{\mathrm{disp}}]$, where \(v_{\mathrm{disp}}\) is the spatial drift vector. These summaries are concatenated with normalized absolute time, inter-event time, absolute location, and location displacement to form the final node representation.

% Placeholder for Figure 1

\subsection{Dual-Stream Spatio-Temporal Encoder}

Temporal and spatial dependencies enter the model through separate streams. The temporal stream handles historical evolution in event time, while the spatial stream operates on graph-structured neighborhoods. Treating them symmetrically from the start tends to blur two different sources of uncertainty: when the next event occurs and where it appears. The encoder therefore maintains distinct pathways before merging them into a shared context.

For event \(i\), the temporal embedding is written as:
\begin{equation}
e_t^{(i)}=
\left[
\frac{\Delta t_i-\mu_{\Delta t}}{\sigma_{\Delta t}}
\;\|\;
\log(1+\exp(\Delta t_i))
\;\|\;
\sin(\omega \Delta t_i)
\;\|\;
\cos(\omega \Delta t_i)
\right].
\end{equation}
This embedding mixes normalized scale, a smooth positive transform, and harmonic components. The first term preserves the relative magnitude of inter-event intervals; the second provides a stable monotone transform for positive temporal variation; the sinusoidal terms make periodic structure easier to express. The temporal sequence is encoded by a Transformer with Rotary Positional Embedding (RoPE) \cite{su2021roformer}:
\begin{equation}
\mathrm{Attention}(Q,K,V)
=
\mathrm{Softmax}\!\left(
\frac{(Q R_{\Theta})(K R_{\Theta})^{\top}}{\sqrt{d_k}}
\right)V.
\end{equation}
RoPE is used here to strengthen relative order information without introducing a separate positional tensor, which is helpful when long-range temporal history is important.

The spatial stream starts from a Fourier lifting of the coordinates, given by $\gamma(s_i)=[\cos(2\pi Bs_i)\;\|\;\sin(2\pi Bs_i)]$, where \(B\) is a sampled frequency matrix \cite{tancik2020fourier}. This mapping enlarges the representational space before graph propagation, making localized spatial variation easier to model. The lifted features are propagated over \(\mathcal{G}\) using stacked GATv2 layers \cite{brody2022how}. The attention coefficient on edge \((j,i)\) is modulated by the edge attribute:
\begin{equation}
\alpha_{ij}=
\frac{
\exp\!\left(
\mathrm{LeakyReLU}\!\left(
a^{\top}[Wh_{i,\mathrm{spat}}^{(l)}\|Wh_{j,\mathrm{spat}}^{(l)}\|W_e e_{ji}]
\right)
\right)
}{
\sum_{k\in \mathcal{N}_i}
\exp\!\left(
\mathrm{LeakyReLU}\!\left(
a^{\top}[Wh_{i,\mathrm{spat}}^{(l)}\|Wh_{k,\mathrm{spat}}^{(l)}\|W_e e_{ki}]
\right)
\right)
},
\end{equation}
and the spatial state is updated as:
\begin{equation}
h_{i,\mathrm{spat}}^{(l+1)}
=
\sigma\!\left(
\sum_{j\in \mathcal{N}_i}\alpha_{ij}W_v h_{j,\mathrm{spat}}^{(l)}
\right).
\end{equation}
Injecting \(e_{ji}\) into the attention coefficient keeps the aggregation graph-aware in a stronger sense: the model does not only learn which neighbors to attend to, but also how their influence should vary across temporal and spatial scales. This is particularly useful when nearby historical events act as local triggers rather than as generic context.

The two streams are fused through a gated interaction layer:
\begin{equation}
C_i
=
\sigma\!\left(
W_{\mathrm{gate}}[h_{i,\mathrm{temp}}\|h_{i,\mathrm{spat}}]
\right)
\odot
\mathrm{MLP}_{\mathrm{fuse}}\!\left(
[h_{i,\mathrm{temp}}\|h_{i,\mathrm{spat}}]
\right),
\end{equation}
which yields the context matrix \(C\in\mathbb{R}^{L\times d}\). A plain concatenation followed by a linear projection is possible, but the gate is useful in practice because the relative importance of temporal and spatial evidence varies across datasets and across histories. The fused context is passed to the reverse diffusion model.

\subsection{DiT-Based Diffusion Estimation}

The prediction target is the next-event variable \(x_0=[\Delta t_{N+1},s_{N+1}]\), and the objective is to learn its conditional distribution given the history. The reverse process is parameterized by a conditional Diffusion Transformer (DiT) \cite{peebles2023scalable}.

Conditioning is injected through Adaptive Layer Normalization. For a hidden state \(h\) in a DiT block, the context-dependent modulation parameters are:
\begin{equation}
[\gamma_1,\beta_1,\alpha_1,\gamma_2,\beta_2,\alpha_2]
=
W_{\mathrm{ada}}e_{\mathrm{cond}}+b_{\mathrm{ada}},
\end{equation}
where \(e_{\mathrm{cond}}=\mathrm{MLP}(C,k)\) depends on both the context and the diffusion step \(k\). The block update takes the form:
\begin{equation}
\tilde{h}
=
h+\alpha_1\odot
\mathrm{SelfAttn}\!\left(
(1+\gamma_1)\odot \mathrm{LayerNorm}(h)+\beta_1
\right),
\end{equation}
\begin{equation}
h^{(l+1)}
=
\tilde{h}+\alpha_2\odot
\mathrm{FFN}\!\left(
(1+\gamma_2)\odot \mathrm{LayerNorm}(\tilde{h})+\beta_2
\right).
\end{equation}
The residual gates are zero-initialized, following standard DiT practice.

A single denoising stream is not used. Temporal and spatial targets are handled by two parallel branches, one for temporal denoising and the other for spatial denoising. This separation matters because the uncertainty over \(\Delta t_{N+1}\) is not of the same type as the uncertainty over \(s_{N+1}\). Temporal prediction is often dominated by historical progression and local interval structure, while spatial prediction tends to be more explicitly multi-modal. A fully shared denoising path makes it harder to preserve these different inductive biases.

Early layers keep the two branches separate. Cross-branch interaction is introduced only after a designated stage. For the temporal branch, the interaction gate is:
\begin{equation}
g_{s\rightarrow t}
=
\sigma\!\left(
W_{\mathrm{gate}}[h_{\mathrm{temp}}^{(l)}\|h_{\mathrm{spat}}^{(l)}]
\right),
\end{equation}
and the updated state is:
\begin{equation}
\hat{h}_{\mathrm{temp}}^{(l)}
=
h_{\mathrm{temp}}^{(l)}+g_{s\rightarrow t}\odot h_{\mathrm{spat}}^{(l)}.
\end{equation}
A symmetric update is applied to the spatial branch. Delaying this interaction is deliberate. If the two branches are mixed too early, branch-specific denoising patterns become weaker and the model tends to over-share features. Restricting interaction to deeper layers keeps the early dynamics specialized while still allowing explicit spatio-temporal coupling when the denoising state becomes more informative.

\subsection{Leap Inference with Prior Guidance}
\label{sec:leap_inference}

Standard reverse diffusion starts from pure Gaussian noise and traverses the full trajectory. In sparse continuous spatio-temporal domains, this global search is expensive and often poorly localized. GLIDE shortens this process by initializing the reverse chain around a deterministic anchor.

\begin{algorithm}[H]
\caption{GLIDE Inference with Leap Sampling}
\label{alg:leap_inference}
\textbf{Input}: Context embeddings $C$, Leap step $\tau$, Guidance scale $w$\\
\textbf{Output}: Predicted next event $x_0$
\begin{algorithmic}[1]
\STATE $\hat{x}_0 \leftarrow \text{JointMeanPredictor}(C)$ \hfill \COMMENT{\textit{Deterministic prior anchor}}
\STATE $\epsilon \sim \mathcal{N}(0, I)$
\STATE $x_{\tau} \leftarrow \sqrt{\bar{\alpha}_{\tau}}\hat{x}_0 + \sqrt{1-\bar{\alpha}_{\tau}}\epsilon$ \hfill \COMMENT{\textit{Leap initialization}}
\FOR{$t = \tau, \tau-1, \dots, 1$}
    \STATE $x_{t-1}^{diff} \leftarrow \text{DiT\_Denoise}(x_t, C, t)$ \hfill \COMMENT{\textit{Standard reverse step}}
    \IF{$t > \tau / 2$}
        \STATE $g_t \leftarrow w \cdot (t / \tau)$
        \STATE $x_{t-1} \leftarrow (1 - g_t) x_{t-1}^{diff} + g_t \hat{x}_0$ \hfill \COMMENT{\textit{Elastic prior guidance}}
    \ELSE
        \STATE $x_{t-1} \leftarrow x_{t-1}^{diff}$
    \ENDIF
\ENDFOR
\STATE \textbf{return} $x_0$
\end{algorithmic}
\end{algorithm}

The anchor is produced by a joint mean predictor trained alongside the diffusion backbone. Given the context matrix \(C\), the predictor outputs:
\begin{equation}
(\hat{t}_0,\hat{s}_0)=\mathrm{JMP}(C),
\qquad
\hat{x}_0=[\hat{t}_0,\hat{s}_0],
\end{equation}
where \(\mathrm{JMP}(\cdot)\) denotes a lightweight joint mean predictor operating on the conditioning context and producing coarse temporal and spatial estimates of the next event. It produces a coarse estimate \(\hat{x}_0\) of the next event, which serves as a deterministic anchor for subsequent refinement. The training objective combines diffusion loss and auxiliary regression loss:
\begin{equation}
L_{\mathrm{total}}
=
(L_{\mathrm{diff}}^t + 2L_{\mathrm{diff}}^s)
+
0.1(L_{\mathrm{mean}}^t + 2L_{\mathrm{mean}}^s).
\end{equation}
The auxiliary regression term is kept small relative to the diffusion loss, so early inaccuracies in \(\hat{x}_0\) do not dominate optimization. In practice, EMA further stabilizes the joint training dynamics.

At inference time, the reverse process starts from an intermediate step \(\tau\) rather than from step \(K\). In practice, \(\tau\) is controlled by a fixed start ratio with respect to the full diffusion horizon and is treated as a hyperparameter selected on the validation set:
\begin{equation}
x_{\tau}
=
\sqrt{\bar{\alpha}_{\tau}}\hat{x}_0
+
\sqrt{1-\bar{\alpha}_{\tau}}\epsilon,
\qquad
\epsilon\sim \mathcal{N}(0,I).
\end{equation}
The sample is thus injected near a high-probability region implied by the history, bypassing a substantial part of the unguided early trajectory. Gaussian noise is retained, so different samples can still explore different local modes around the anchor instead of collapsing to a single deterministic point. Algorithm~1 summarizes the procedure.

The remaining reverse updates are further stabilized by an elastic anchor guidance rule:
\begin{equation}
x_{t-1}
\leftarrow
(1-g_t)x_{t-1}^{\mathrm{diff}} + g_t\hat{x}_0,
\qquad
g_t = w\cdot \frac{t}{\tau}.
\end{equation}
The guidance weight decays linearly as \(t\) approaches zero, so the trajectory moves from coarse localization to fine stochastic refinement.

\section{Experiments}

We evaluate GLIDE on three primary real-world benchmarks: \textit{Earthquake}, \textit{COVID-19}, \textit{Citibike}, and include additional analysis on \textit{Crime} for graph-related ablations. The preprocessing of \textit{Earthquake}, \textit{COVID-19}, and \textit{Citibike} follows DSTPP \cite{yuan2023dstpp}: Earthquake is segmented by a 30-day sliding window with a 7-day gap, COVID-19 is segmented by a 7-day window with a 3-day gap, and Citibike is split into one-day subsequences for each bike starting from 5:00 a.m. The \textit{Crime} dataset follows the split used in SMASH \cite{li2024beyond}; unlike the original marked setting, we remove the category dimension and retain only temporal and spatial coordinates.

\subsection{Experimental Setup}

All datasets are split chronologically into training, validation, and test sets with a ratio of 8:1:1. The comparison includes classical intensity-based models and recent neural or generative baselines: Hawkes \cite{hawkes1971spectra}, RMTPP \cite{du2016recurrent}, SAHP \cite{zhang2020self}, NJSDE \cite{jia2019njsde}, NSTPP \cite{chen2021nstpp}, DeepSTPP \cite{zhou2022deepstpp}, DSTPP \cite{yuan2023dstpp}, and SMASH \cite{li2024beyond}. These baselines span recurrent and self-attentive neural point processes, continuous-time generative models, and recent diffusion- or score-based approaches.

Distribution fitting is evaluated by per-event negative log-likelihood (NLL), reported separately for spatial and temporal components. For next-event prediction, spatial error is measured by Euclidean distance and temporal error by RMSE. Lower values indicate better performance for all metrics.

GLIDE is trained for 1000 epochs with batch size 64. The forward diffusion process uses 500 steps. During sampling, we adopt DDIM-style accelerated sampling \cite{song2021ddim} and use 50 reverse steps unless otherwise stated. During evaluation, 20 samples are drawn for each conditioning context and averaged to obtain the reported metrics. Hidden dimension, dropout, and leap-related hyperparameters are selected on the validation set.The multi-scale window sets are fixed per dataset: \(\{64,32,16,8\}\) for Earthquake, \(\{14,7\}\) for COVID-19, temporal \(\{32,16,8,4\}\) and spatial \(\{8,6,4,2\}\) for Citibike, and \(\{32,16,8,4\}\) for Crime.For leap inference, the start ratio and guidance scale are selected on the validation set and then fixed across the three primary benchmark datasets. In the final configuration, we use a start ratio of 0.4 and a guidance scale of 0.1 for Earthquake, COVID-19, and Citibike. Therefore, the leap step is $\tau = \lfloor 0.4T \rfloor$.

\begin{table}[!t]
    \centering
    \small
    \renewcommand{\arraystretch}{0.95}
    
    % ================= Table 1: NLL =================
    \caption{Performance evaluation for negative log-likelihood per event on test data. $\downarrow$ means lower is better. \textbf{Bold} denotes the best results. }
    \label{tab:performance_nll}
    
    \resizebox{\textwidth}{!}{%
    \begin{tabular}{lcccccc} 
        \toprule
        \multirow{2}{*}{Model} & \multicolumn{2}{c}{Earthquake} & \multicolumn{2}{c}{COVID-19} & \multicolumn{2}{c}{Citibike} \\
        \cmidrule(lr){2-3} \cmidrule(lr){4-5} \cmidrule(lr){6-7} 
         & Spatial $\downarrow$ & Temporal $\downarrow$ & Spatial $\downarrow$ & Temporal $\downarrow$ & Spatial $\downarrow$ & Temporal $\downarrow$ \\
        \midrule
        % 经典但有缺失的模型
        CNF & $1.35_{\pm 0.000}$ & - & $2.05_{\pm 0.014}$ & - & $2.15_{\pm 0.000}$ & - \\
        Hawkes & - & $-0.514_{\pm 0.000}$ & - & $-2.06_{\pm 0.000}$ & - & $-1.06_{\pm 0.001}$ \\
        RMTPP & - & $0.0930_{\pm 0.051}$ & - & $-1.30_{\pm 0.022}$ & - & $1.24_{\pm 0.001}$ \\
        SAHP & - & $-0.229_{\pm 0.007}$ & - & $-1.37_{\pm 0.118}$ & - & $-1.02_{\pm 0.067}$ \\
        \midrule
        % 完整数据的现代模型
        NJSDE & $1.65_{\pm 0.012}$ & $0.0950_{\pm 0.203}$ & $2.21_{\pm 0.005}$ & $-1.82_{\pm 0.002}$ & $2.63_{\pm 0.001}$ & $-0.804_{\pm 0.059}$ \\
        NSTPP & $0.885_{\pm 0.037}$ & $-0.623_{\pm 0.004}$ & $1.90_{\pm 0.017}$ & $-2.25_{\pm 0.002}$ & $2.38_{\pm 0.053}$ & $-1.09_{\pm 0.004}$ \\
        DeepSTPP & $4.92_{\pm 0.007}$ & $-0.174_{\pm 0.001}$ & $\mathbf{0.361}_{\pm 0.01}$ & $-1.09_{\pm 0.01}$ &$\mathbf{ -4.94}_{\pm 0.016}$ & $-1.13_{\pm 0.002}$ \\
        DSTPP  & $0.452_{\pm 0.004}$ & $-1.05_{\pm 0.014}$ & $0.500_{\pm 0.017}$ & $-2.44_{\pm 0.005}$ & $0.515_{\pm 0.013}$ & $-2.41_{\pm 0.010}$  \\
        SMASH  & $0.376_{\pm 0.01}$ & $-1.231_{\pm 0.001}$ & $0.482_{\pm 0.002}$ & $-0.02_{\pm 0.006}$ & $0.232_{\pm 0.001}$ & $-2.34_{\pm 0.002}$ \\
        \midrule
        GLIDE(Ours)  & $-\mathbf{0.067}_{\pm 0.03}$ & $-\mathbf{1.299}_{\pm 0.05}$ & $0.3739_{\pm 0.02}$ & $-\mathbf{2.476}_{\pm 0.020}$ & ${0.373}_{\pm 0.07}$ & $-\mathbf{2.44}_{\pm 0.010}$ \\
        \bottomrule
    \end{tabular}%
    }

    \vspace{0.1cm}

    \vspace{0.6cm} % 两个表格之间的垂直间距
    
    % ================= Table 2: RMSE & Distance =================
    \caption{Performance evaluation for predicting both time and space of the next event. We use Euclidean distance to measure the prediction error of the spatial domain and RMSE for time prediction.}
    \label{tab:performance_rmse}
    
    \resizebox{\textwidth}{!}{%
    \begin{tabular}{lcccccc} 
        \toprule
        \multirow{2}{*}{Model} & \multicolumn{2}{c}{Earthquake} & \multicolumn{2}{c}{COVID-19} & \multicolumn{2}{c}{Citibike} \\
        \cmidrule(lr){2-3} \cmidrule(lr){4-5} \cmidrule(lr){6-7} 
         & Spatial $\downarrow$ & Temporal $\downarrow$ & Spatial $\downarrow$ & Temporal $\downarrow$ & Spatial $\downarrow$ & Temporal $\downarrow$ \\
        \midrule
        % 经典但有缺失的模型
        CNF & $8.48_{\pm 0.054}$ & - & $0.559_{\pm 0.000}$ & - & $0.722_{\pm 0.000}$ & - \\
        Hawkes & - & $0.544_{\pm 0.010}$ & - & $0.672_{\pm 0.088}$ & - & $0.534_{\pm 0.011}$ \\
        RMTPP & - & $0.424_{\pm 0.009}$ & - & $1.32_{\pm 0.024}$ & - & $2.07_{\pm 0.015}$ \\
        SAHP & - & $0.409_{\pm 0.002}$ & - & $0.184_{\pm 0.024}$ & - & $0.203_{\pm 0.010}$ \\
        \midrule
        % 完整数据的现代模型
        NJSDE & $9.98_{\pm 0.024}$ & $0.465_{\pm 0.009}$ & $0.641_{\pm 0.009}$ & $0.137_{\pm 0.001}$ & $0.707_{\pm 0.001}$ & $0.264_{\pm 0.005}$ \\
        NSTPP & $8.11_{\pm 0.000}$ & $0.547_{\pm 0.010}$ & $0.560_{\pm 0.000}$ & $0.145_{\pm 0.002}$ & $0.705_{\pm 0.000}$ & $0.355_{\pm 0.013}$ \\
        DeepSTPP & $9.20_{\pm 0.000}$ & $\mathbf{0.341}_{\pm 0.000}$ & $0.687_{\pm 0.000}$ & $0.197_{\pm 0.000}$ & $0.044_{\pm 0.000}$ & $0.234_{\pm 0.000}$ \\
        DSTPP  & $7.13_{\pm 0.112}$ & $0.379_{\pm 0.001}$ & $0.432_{\pm 0.001}$ & $0.096_{\pm 0.001}$ & $0.032_{\pm 0.000}$ & $0.207_{\pm 0.001}$ \\
        SMASH  & $8.04_{\pm 0.221}$ & $0.380_{\pm 0.005}$ & $0.563_{\pm 0.001}$ & $0.100_{\pm 0.007}$ & $\mathbf{0.031}_{\pm 0.001}$ & $\mathbf{0.204}_{\pm 0.002}$ \\
        \midrule
        GLIDE(Ours)  & $\mathbf{6.61}_{\pm 0.150}$ & $0.378_{\pm 0.001}$ & $\mathbf{0.394}_{\pm 0.005}$ & $\mathbf{0.095}_{\pm 0.000}$ & $\mathbf{0.031}_{\pm 0.001}$ & $0.207_{\pm 0.002}$ \\
        \bottomrule
    \end{tabular}%
    }
\end{table}

\subsection{Main Results}

Table~\ref{tab:performance_nll} reports distribution fitting results, and Table~\ref{tab:performance_rmse} reports next-event prediction accuracy. Across the three primary benchmark datasets, the gain is strongest on the spatial side, while temporal accuracy remains competitive.

The NLL improvement is most pronounced on \textit{Earthquake}, where the spatial NLL drops from \(0.452\) for DSTPP to \(-0.067\). This margin is notable because \textit{Earthquake} is also the sparsest dataset in the benchmark. In such settings, standard diffusion is more likely to disperse probability mass across implausible regions before the reverse chain contracts to a meaningful support. The graph-conditioned context and prior-guided initialization in GLIDE reduce exactly this effect. The same trend appears on \textit{COVID-19} and \textit{Citibike}, where the spatial NLL decreases from \(0.500\) to \(0.3739\) and from \(0.515\) to \(0.373\), respectively. The improvement therefore does not depend on one particular type of spatio-temporal dynamics.

The temporal NLL results are more moderate, which is expected. Compared with spatial generation, the temporal component is typically lower-dimensional and less explicitly multi-modal, so strong gains are harder to obtain. What matters here is that the spatial improvement does not come with a temporal penalty. GLIDE maintains competitive temporal likelihood while substantially improving spatial density estimation, which is precisely the regime where deterministic or weakly-structured models tend to struggle most.

The next-event prediction results show a similar advantage. GLIDE attains the lowest spatial error on \textit{Earthquake} and \textit{COVID-19}, and matches the best result on \textit{Citibike}. On \textit{Earthquake}, the spatial error falls to \(6.61\) km, compared with \(9.20\) km for DeepSTPP and \(8.04\) km for SMASH. Temporal RMSE remains competitive across all three datasets. The spatial gains are therefore not obtained by sacrificing the temporal side. A more plausible interpretation is that the deterministic anchor gives the model a better starting region, after which diffusion is still free to perform stochastic refinement around plausible modes. This is also consistent with the improvement in spatial NLL: the model is not only predicting better points, but also assigning probability mass more coherently around the true support.

\subsection{Ablation and Mechanism Analysis}

Three ablations isolate the main components of GLIDE: \textit{w/o Leap}, which removes prior-guided leap initialization and falls back to standard DDIM sampling \cite{song2021ddim} from pure noise; \textit{w/o Interaction}, which removes cross-stream gating; and \textit{w/o Graph}, which removes spatial message passing.

\begin{figure}[t]
    \centering
    % 左侧图 3
    \begin{minipage}[b]{0.48\textwidth}
        \centering
        \includegraphics[width=\textwidth]{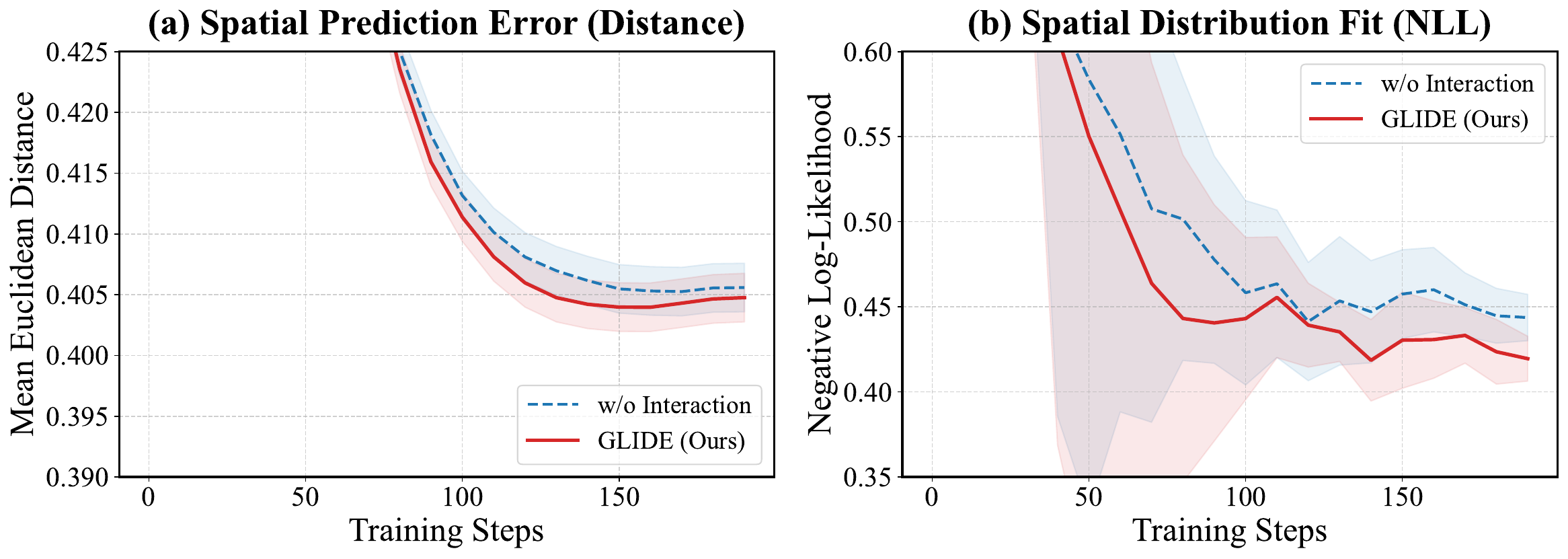}
        \caption{Ablation on Spatio-Temporal Interaction.}
        \label{fig:interaction}
    \end{minipage}
    \hfill % 在两张图中间插入弹性空白
    % 右侧图 4
    \begin{minipage}[b]{0.48\textwidth}
        \centering
        \includegraphics[width=\textwidth]{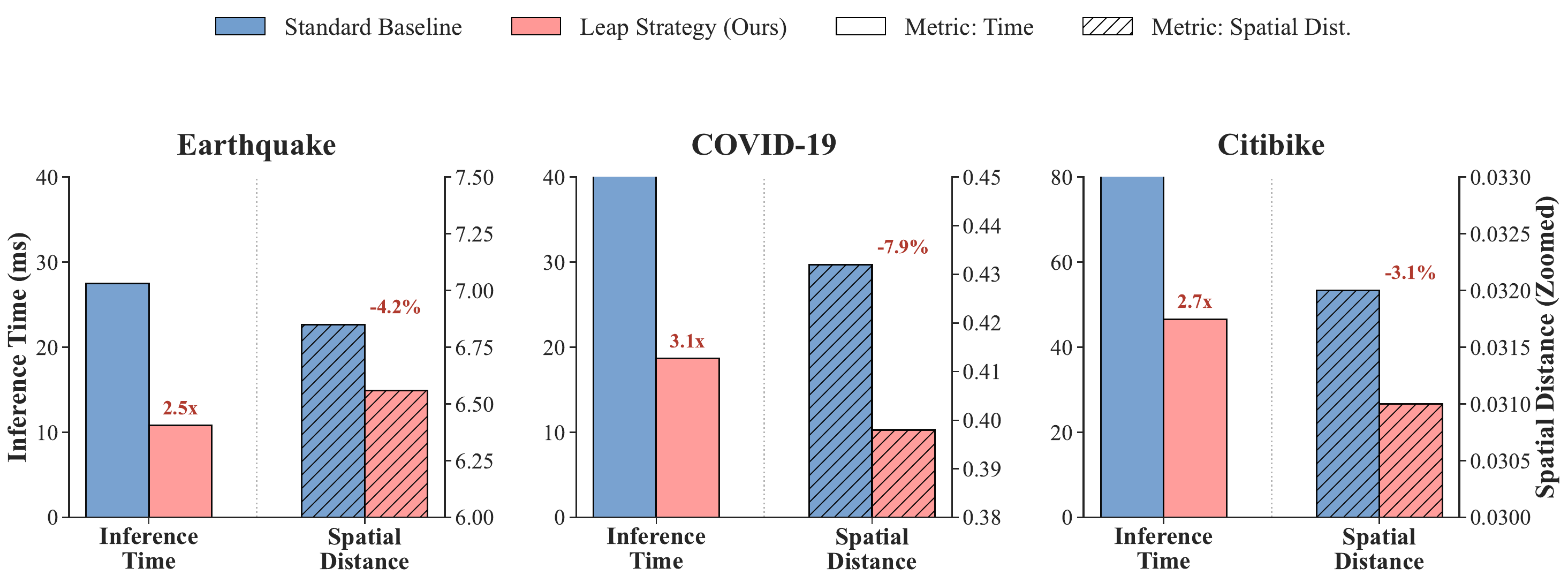}
        \caption{Ablation study on sampling strategy.}
        \label{fig:leap}
    \end{minipage}
\end{figure}

\paragraph{Cross-stream interaction.}
Figure~\ref{fig:interaction} shows that removing the interaction module affects spatial learning more than temporal learning. On \textit{COVID-19}, the spatial NLL becomes less stable once the temporal branch no longer informs the spatial branch. This is consistent with the role of temporal phase information in spatial generation: without it, the spatial branch has more difficulty concentrating probability mass on the appropriate regions.

If the two branches were largely interchangeable, removing cross-stream gating would have only a minor impact. Instead, the degradation is systematic on the spatial side, suggesting that temporal information is not easily recoverable from spatial context alone.

\begin{figure}[t]
    \centering
    \vspace{-0.3cm} % 吞掉图表上方的白边
    % 将宽度从 \textwidth 缩小到 0.75 或 0.8
    \includegraphics[width=0.75\textwidth]{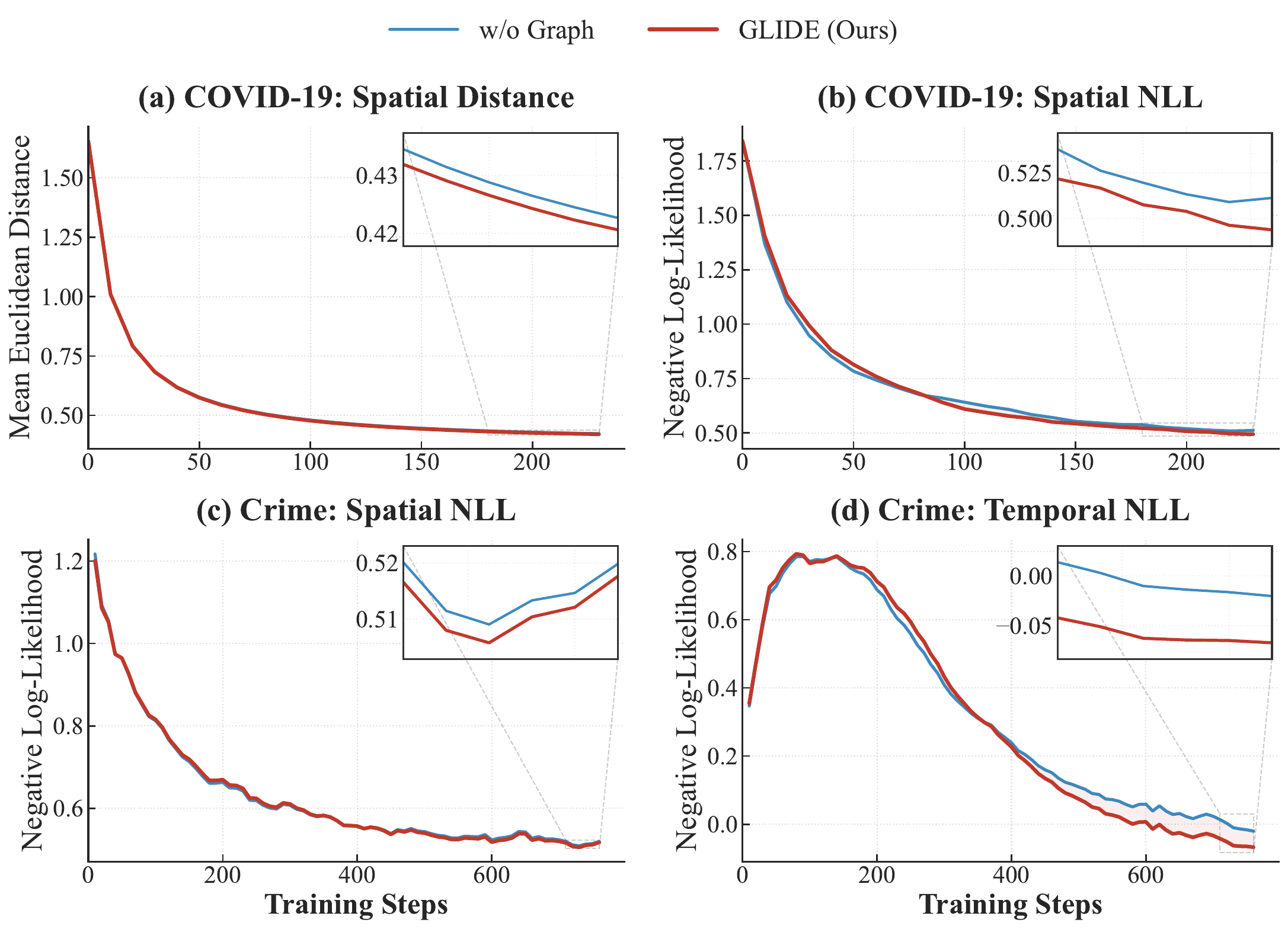} 
    \vspace{-0.3cm} % 吞掉图片和标题之间的白边
   \caption{Ablation of graph structure on the COVID-19 and Crime datasets. Removing graph guidance leads to worse NLL in contagion-driven settings.}
    \label{fig:graph}
    \vspace{-0.4cm} % 吞掉标题下方与正文之间的白边
\end{figure}
\paragraph{Leap initialization.}
Figure~\ref{fig:leap} compares prior-guided leap sampling with standard DDIM sampling. In the baseline setting, reverse generation starts from pure noise. In GLIDE, sampling starts from a prior-guided intermediate state and uses 50 reverse steps. The wall-clock speedup reaches \(2.5\times\), \(3.1\times\), and \(2.7\times\) on \textit{Earthquake}, \textit{COVID-19}, and \textit{Citibike}, respectively. The shorter and better-initialized reverse process does not reduce prediction quality; spatial error is reduced by \(6.0\%\), \(7.9\%\), and \(3.1\%\) on the same datasets.

The result suggests that a substantial part of standard diffusion sampling is spent on an early search phase in which the reverse chain is still far from the relevant support. Leap initialization removes much of this phase by placing the sample near a plausible high-density region before refinement begins. The gain is therefore not only a matter of using fewer effective reverse updates, but also of avoiding the least informative part of the trajectory.

\paragraph{Graph guidance.}
Figure~\ref{fig:graph} clarifies the effect of the graph module. The gain is strongest on \textit{COVID-19} and \textit{Crime}, where local propagation and neighborhood structure are central to the dynamics. Removing the graph in these settings leads to clear degradation in NLL. The effect is smaller on \textit{Earthquake} and \textit{Citibike}, suggesting that graph guidance is most effective when nearby historical events provide direct structural information about the next event.

The graph is therefore not a uniformly strong prior across all tasks. Its benefit is largest when local interaction is part of the underlying process.

\begin{figure*}[t]
    \centering
    
    % --- 第一张子图 (a) Earthquake ---
    \begin{subfigure}{\textwidth}
        \centering
        % 替换为你的实际文件名
        \includegraphics[width=\linewidth]{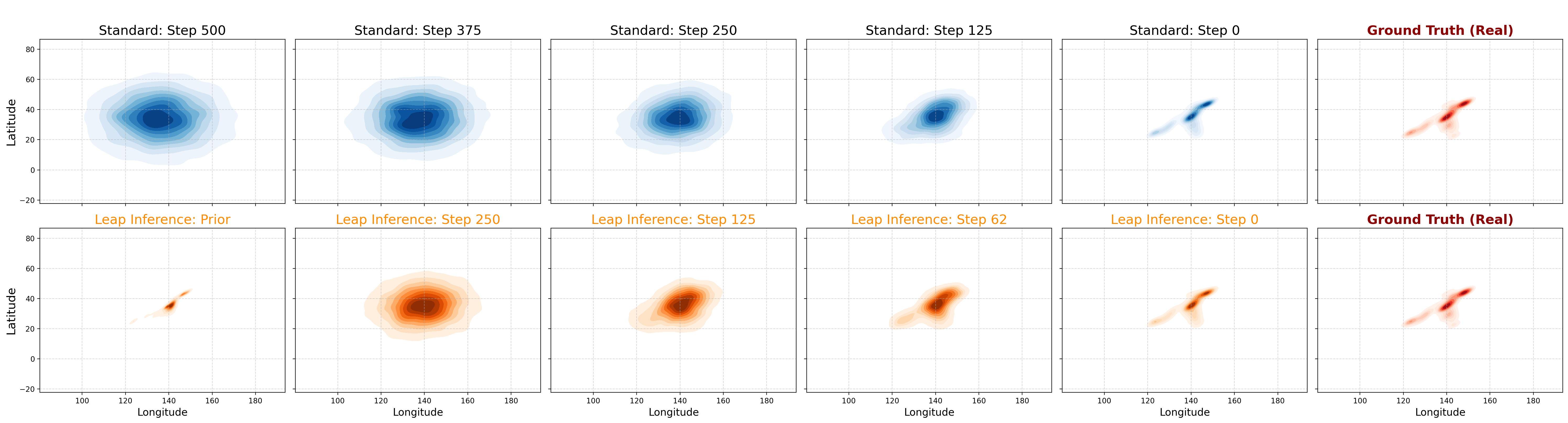}
        \caption{Generative trajectories on the \textbf{Earthquake} dataset.}
        \label{fig:traj_earthquake}
    \end{subfigure}
    
    \vspace{0.3cm} % 两张图之间的垂直间距，可自行微调
    
    % --- 第二张子图 (b) COVID-19 ---
    \begin{subfigure}{\textwidth}
        \centering
        % 替换为你的实际文件名
        \includegraphics[width=\linewidth]{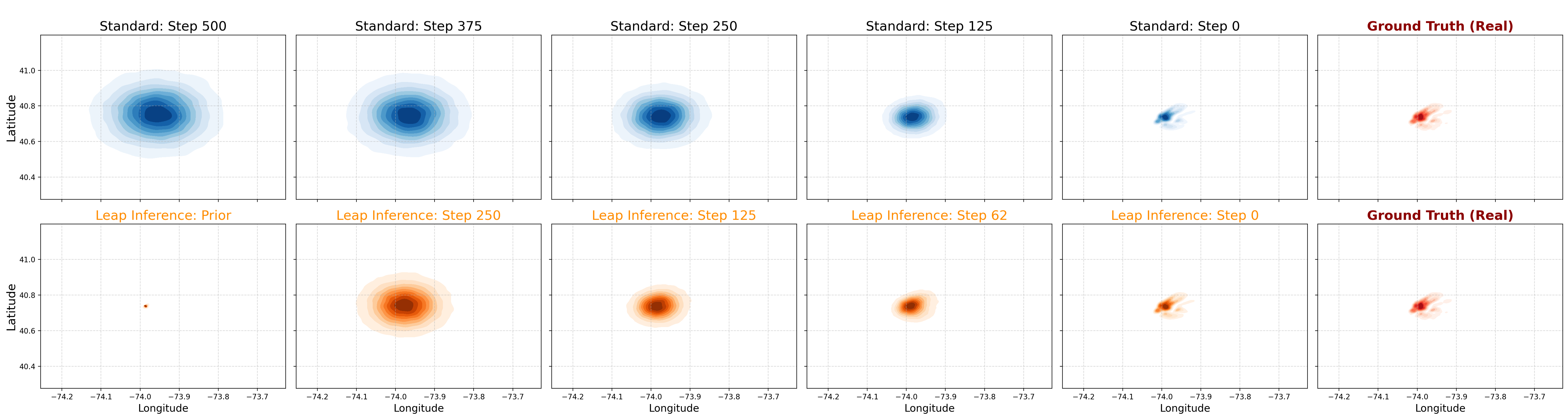}
        \caption{Generative trajectories on the \textbf{Citibike} dataset.}
        \label{fig:traj_covid}
    \end{subfigure}
    
    % --- 整个大图的总图注 ---
    \caption{Generative trajectories (KDE). Standard diffusion (top) remains spatially diffuse for longer, whereas leap inference (bottom) starts from a prior-guided state and reaches high-density regions earlier.}
    \label{fig:trajectories}
\end{figure*}

\subsection{Qualitative Analysis of Generative Trajectories}

Figure~\ref{fig:trajectories} visualizes the reverse trajectories on \textit{Earthquake} and \textit{Citibike}. Standard diffusion remains spatially diffuse for a large portion of the reverse chain and only gradually contracts toward plausible modes. GLIDE enters high-density regions much earlier because the trajectory starts from a prior-guided intermediate state rather than from pure isotropic noise. The later denoising steps are then used mainly for local refinement instead of global search.

The contrast is particularly informative because it appears in both sparse and dense regimes. On \textit{Earthquake}, the difference is visible in how quickly the samples stop exploring empty regions. On \textit{Citibike}, the gap is less about empty space and more about how early the trajectory begins to concentrate around realistic mobility hubs. The visual pattern therefore matches the quantitative results from two angles at once: lower NLL reflects better concentration of probability mass, and lower inference time reflects the removal of an extended coarse-search phase.

\subsection{Efficiency Analysis}

Table~\ref{tab:efficiency} highlights the parameter efficiency of the leap mechanism. In both DSTPP and GLIDE, the additional parameters introduced by Leap are very small compared with the full backbone. For example, DSTPP+Leap adds only 0.0257M parameters on top of 1.5844M, and GLIDE adds 0.1685M leap-related parameters on top of 2.6920M. Despite this small overhead, Leap consistently improves efficiency, reducing the average generation time per event from 0.045 ms to 0.023 ms in DSTPP and from 0.275 ms to 0.109 ms in GLIDE. This suggests that the gain brought by Leap is highly cost-effective: a lightweight module is sufficient to produce a clear improvement in inference efficiency.
\begin{table}[H]
    \centering
    \scriptsize % 进一步缩小字体到 scriptsize
    \renewcommand{\arraystretch}{1.0} 
    \setlength{\tabcolsep}{4pt} % 极限压缩列边距
    \caption{Efficiency analysis of Leap. Time denotes the average latency per generated event.}
    \label{tab:efficiency}
    \begin{tabular}{lccc} 
        \toprule
        \multirow{2}{*}{\textbf{Model}} & \textbf{Params} & \textbf{Leap Params} & \textbf{Time} \\
         & \textbf{(M)} & \textbf{(M)} & \textbf{(ms) $\downarrow$} \\
        \midrule
        DSTPP & 1.5844 & - & 0.045 \\
        DSTPP + Leap & 1.6101 & 0.0257 & 0.023 \\
        \midrule
        GLIDE (w/o Leap) & 2.6920 & - & 0.275 \\
        GLIDE (Ours) & 2.8605 & 0.1685 & 0.109 \\
        \bottomrule
    \end{tabular}
\end{table}

\section{Conclusion}

GLIDE combines graph-structured historical encoding with conditional diffusion for spatio-temporal point process modeling. A multi-scale historical graph captures local historical structure, a dual-stream encoder separates temporal evolution from spatial topology, and a dual-branch Diffusion Transformer models the next-event distribution. Leap inference further shortens the reverse trajectory by starting from a prior-guided intermediate state rather than from pure noise. Experiments on multiple real-world datasets show consistent gains in both distribution fitting and next-event prediction, with the clearest improvements appearing on the spatial side. The results suggest that GLIDE improves not only point accuracy but also the allocation of probability mass in space, while reducing reverse-sampling cost. The current formulation is limited to continuous temporal and spatial coordinates and does not explicitly model event marks or categorical types, and the graph component is most effective when local interaction is part of the underlying process. Extending GLIDE to marked STPPs, dynamically evolving graph structures, and adaptive leap-step selection would be natural directions for future work.

\bibliographystyle{splncs04}
\bibliography{wenxian} 
\end{document}